\documentclass[letterpaper, 10 pt, conference]{ieeeconf}  

\IEEEoverridecommandlockouts                              

\usepackage{caption}
\usepackage[T1]{fontenc}
\usepackage{amsfonts}
\usepackage[cmex10]{amsmath}
\DeclareMathOperator*{\E}{\mathbb{E}}
\usepackage[hidelinks]{hyperref}
\usepackage{graphicx}
\usepackage{algorithm}%
\usepackage{algpseudocode}%
\usepackage{todonotes}

\title{\LARGE \bf
Real-World Human-Robot Collaborative Reinforcement Learning*
}

\author{Ali Shafti$^{1}$, Jonas Tjomsland$^{1}$, William Dudley$^{1}$ and A. Aldo Faisal$^{1}$
\thanks{$^{1}$ AS, JT, WD and AAF. are with the Brain and Behaviour Lab, Dept. of Bioengineering and Dept. of Computing, Imperial College London, SW7 2AZ, London, UK {\tt\small \{a.shafti,a.faisal\}@imperial.ac.uk}}%
}

\begin{document}

\maketitle
\thispagestyle{empty}
\pagestyle{empty}

\begin{abstract}

The intuitive collaboration of humans and intelligent robots (embodied AI) in the real-world is an essential objective for many desirable applications of robotics. Whilst there is much research regarding explicit communication, we focus on how humans and robots interact implicitly, on motor adaptation level. We present a real-world setup of a human-robot collaborative maze game, designed to be non-trivial and only solvable through collaboration, by limiting the actions to rotations of two orthogonal axes, and assigning each axes to one player. This results in neither the human nor the agent being able to solve the game on their own. We use deep reinforcement learning for the control of the robotic agent, and achieve results within 30 minutes of real-world play, without any type of pre-training. We then use this setup to perform systematic experiments on human/agent behaviour and adaptation when co-learning a policy for the collaborative game. We present results on how co-policy learning occurs over time between the human and the robotic agent resulting in each participant's agent serving as a representation of how they would play the game. This allows us to relate a person's success when playing with different agents than their own, by comparing the policy of the agent with that of their own agent.

\end{abstract}

\section{INTRODUCTION}
\label{sec:introduction}
Human-Machine Interaction methods are changing. Efforts were previously focused on creating ``user-friendly'' interfaces, so that human users can better learn to work with a system that is persistent in its behaviour. With the ever-increasing success of artificially intelligent agents, however, the possibilities for creating a fluid, adaptive and ever improving interface and interaction are increasing. Instead of the conventional paradigm of the human adapting to the machine, we want machines that can adapt to humans -- a mutual adaptation happening over time, leading to more intuitive and explainable interactions. To achieve this, we need intelligent control agents that can learn as they interact with a human user. A common tool for this is reinforcement learning (RL) as it follows the same learning mechanism driving human learning \cite{wolpert2011principles}. We are interested in implementing Human-in-the-Loop RL, i.e. having an agent that interacts with and learns directly from a human counterpart. 

\begin{figure}[tp]
    \centering
    \includegraphics[width=\columnwidth]{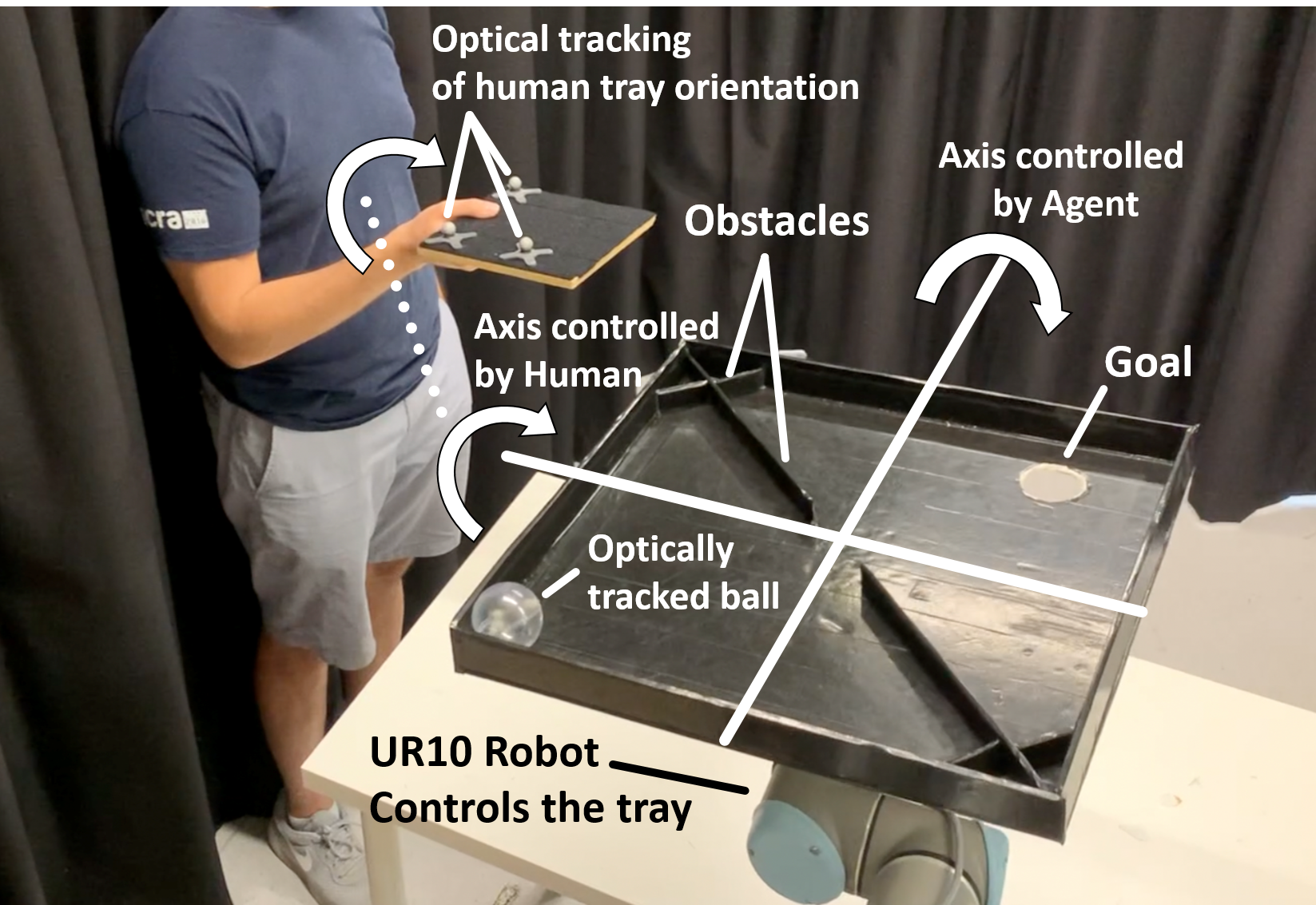}
    \caption{Our Human-Robot co-learning setup: A ball and maze game is designed to require two players for success; one player per rotation axis of the tray. One axis is tele-operated by a human player, and the other axis by a deep reinforcement learning agent. The game can only be solved through collaboration.}
    \label{setup_page1}
\vspace{-5pt}
\end{figure}

Human in-the-loop RL can be mapped as a specific case of a multi-agent system, which has been an ongoing area of research for the past two decades \cite{Stone1997MultiagentSystems,hernandez2019survey}. However, specific complications arise from having a human in-the-loop, mainly due to the stochastic nature of human behaviour, and limited observability of human intent, reasoning and theory of mind. Similarly, the agent is not fully observable for the human (e.g. lack of explainability), causing challenges for interactive learning, making this area unique in its challenges.

Within Robotics, the advent of safe collaborative robots presents an opportunity for use of human in-the-loop systems, i.e. creating robotic systems that learn through collaboration, and develop personalised collaborative policies with humans, through interaction -- similar to how two co-workers would do over time. We present a real-world setup for studies on how humans and intelligent robotic agents can learn and adapt together for the completion of a non-trivial collaborative motor task. We have designed a human-agent collaborative maze game, see Figure \ref{setup_page1}, where a tray needs to be tilted to navigate a ball to a goal. The human controls one axis of tilt, and the agent controls the other. Hence, the agent and the human need to learn to collaborate together. We report the methods used in creating the setup, followed by experiments investigating the possibility, and results of human-robot real-time, real-world collaborative learning.

\section{RELATED WORK}
\label{sec:related}
Human in-the-loop intelligent systems are a topic of active research. In many cases, this takes the form of human-aided learning, e.g. the TAMER framework \cite{Knox2009InteractivelyFramework,Warnell2018DeepSpaces} where the agent learns via real-time qualitative feedback from a human advisor rather than environment reward, outperforming both  humans  and  state-of-the-art  RL  algorithms  in  ATARI Bowling  within  15  minutes. Similarly, sparse  human  feedback  has been shown to result in successful complex behaviour in ATARI and MuJoCo environments within an hour of human time \cite{Christiano2017DeepPreferences}. More active roles for the human counterparts can be seen in cases of shared autonomy. Here, environments  involve multiple  agents (human or artificial) acting at the same time, to achieve shared or  individual  goals. This has been used e.g. in drone-flying tasks, where sub-optimal human actions are augmented by a deep RL agent, to achieve better performance \cite{shared} or in human-robot teaming setups \cite{javdani2018shared}.

An important aspect of such human in-the-loop systems is how to interface the human \cite{shafti2020learning}. We have shown low-cost, intuitive human interfacing through head movements monitoring \cite{headmouse}. Human gaze is an intuitive interface for implicit robot interaction e.g. with assistive robotic systems \cite{shafti2019gaze}, but also to learn and predict human visual attention, e.g. while driving, to train better performing, human interpretable self-driving agents \cite{makrigiorgos2019human}. Other aspects of human behaviour, such as kinematic and muscle activity data can be used e.g. for creating ergonomic, intuitive human-robot collaboration \cite{shafti2019real} or with supervised learning approaches to improve the performance of robotic prosthetics \cite{xiloyannis2017gaussian}.
Deep  RL  has been used to infer social  norms regarding  pedestrian  behaviour to motion plan robotic vehicles in a meaningful manner for humans \cite{Chen2017SociallyLearning}. Implicit interfacing of human advice has been implemented through an agent that modulates its  speed based on how unsure of an action it is, naturally prompting the human for feedback \cite{Macglashan2016ADescriptors}. Opponent modelling and theory of mind have been leveraged to gain insights regarding RL in multi-agent scenarios, showing that in tasks relying on collaboration being aware of the other learning agent, leads to better performance \cite{foerster:aamas18}. 

The above approaches in human in-the-loop systems are either trained in simulation, operate solely in a simulated world or are based on sequential or interval-based interactions involving explicit communication or unbalanced responsibilities within the task. In this work we are  interested instead in real-world, real-time collaborative learning between a human and an agent, with implicit communication.

\section{METHODS}
\label{sec:methods}
\subsection{Robotic Setup}
\label{sec:methods_robot}
\begin{figure}[tp]
    \centering
    \includegraphics[width=\columnwidth]{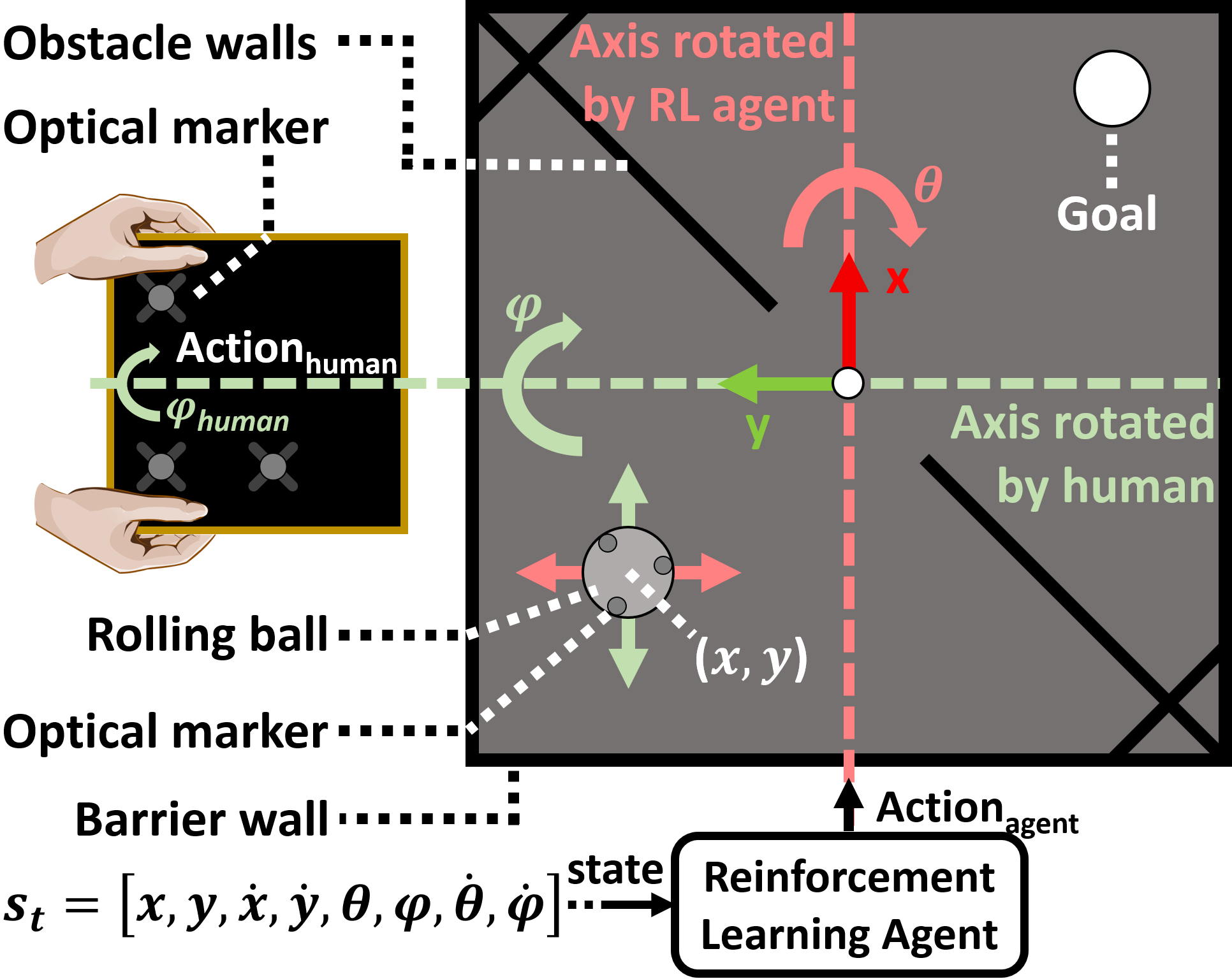}
    \caption{Overview of our collaborative maze game setup. The human and the RL agent's actions are mapped to orthogonal rotation axes of the tray. The individual effects of human and agent actions are marked on the ball with green and red arrows respectively. The states fed to the RL agent are $x$ and $y$ position of the ball in the tray frame, $x$ and $y$ ball velocity, rotation angles along the two axes ($\theta$ for $x$ and $\phi$ for $y$ rotations), and respective rotational velocities.}
    \label{setup_methods}
\vspace{-5pt}
\end{figure}

We use a Universal Robots UR10 (Universal Robots A/S, Odense, Denmark) as the robotic manipulator. A $50\textrm{cm}\times50\textrm{cm}$ square tray is built out of cardboard material, and attached to the UR10 end-effector, through a 3D printed mechanical interface. The tray has barrier walls on all four sides to keep the ball from falling off as well as two obstacle walls, positioned diagonally, with a $9\textrm{cm}$ opening in the centre (refer to Figure \ref{setup_page1} and Figure \ref{setup_methods}). A $5\textrm{cm}$-diameter hole is cut near one of the board's corners, representing the goal for a rolling ball to fall into. The ball is $6\textrm{cm}$ in diameter and made out of transparent acrylic. The game, i.e. the task of rolling the ball from a given start point on the tray, to the goal, is solved purely by rotating the tray around its $x$-$y$ axes (two orthogonal axes on the tray plane, with the centre of the square tray as the origin -- see Figure \ref{setup_methods}); no rotation around the $z$-axis, nor translations along any axes is allowed. The human player's commands are sent via a smaller tray that they hold and rotate (Figure \ref{setup_page1} and \ref{setup_methods}). The human tray's orientation is tracked with three optical markers placed on top of it, through a motion capture system consisting of Optitrack Flex 13 cameras (NaturalPoint, Inc. DBA OptiTrack, Corvallis, Oregon, USA). The position of the ball on the tray is similarly tracked via optical markers placed inside it (Figure \ref{setup_page1} and \ref{setup_methods}).

To integrate the above, we used the Robot Operating System (ROS) \cite{quigley2009ros}, running on a Linux workstation (ROS Melodic, Ubuntu 18.04). The motion capture software was running on a Windows 10 workstation networked with the ROS master. Through ROS, we are able to track the position of the ball with respect to the tray frame: $(x,y)$, the rotation of the tray along its $x$ and $y$ axes: $\theta$ and $\phi$ respectively, and the rotation of the human's handheld tray along its $y$ axis: $\phi_{human}$, see Figure \ref{setup_methods}. Human actions are then calculated through a simple proportional control setup:


\vspace{-10pt}
\begin{equation} \label{eq:1}
    a_{human} = k_p(\phi_{human} - \phi) 
\vspace{-5pt}
\end{equation}

\noindent{with $k_p=2$ selected empirically for our setup and $a_{human}$ limited to $[-1,1]$, i.e. smaller and larger values set to $-1$ and $1$ respectively. To send motion commands to the robot, we used the jog\textunderscore arm\footnote{\href{https://github.com/UTNuclearRoboticsPublic/jog\_arm}{https://github.com/UTNuclearRoboticsPublic/jog\_arm}} ROS package which simplifies the communication of smooth velocity commands to ROS-enabled robots, allowing us to send real-time jogging commands.}

\subsection{Reinforcement Learning Setup}
\label{sec:methods_RL}
Reinforcement learning (RL) agents explore and navigate in an environment, taking actions ($a$) given a state ($s$), with the intention of maximising a long-term reward \cite{RL}. This results in trajectories of state-action transitions, $\tau = (s_{0},a_{0}, s_{1}, a_{1}, ..)$, generated by following a specific policy $\pi$ that provides a mapping from states to actions. For our setup, we define an eight-dimensional state space, 

\vspace{-10pt}
\begin{equation}\label{eq:3}
    s = [x,y,\dot{x},\dot{y},\theta,\phi,\dot{\theta},\dot{\phi}]
\vspace{-5pt}
\end{equation}

\noindent{consisting of the position ($x$, $y$) and velocity ($\dot{x}$, $\dot{y}$) of the ball in the game tray frame, and rotation angles ($\theta$,$\phi$) and rotational velocities ($\dot{\theta}$,$\dot{\phi}$) of the game tray about its $x$ and $y$ axes respectively. Human behaviour is included in the state space through the game tray rotation around its $y$-axis, which is mimicking the human's tray via the tele-operation interface. The RL agent's action space is one-dimensional, a continuous value $a_t\in[-1,1]$ is mapped to rotational velocity commands along the game tray's $x$ axis. We use a sparse reward function,}

\vspace{-10pt}
\begin{equation}\label{eq:4}
  r(s,a)=\begin{cases}
    +10, & \text{goal reached}\\
    -1, & \text{otherwise}
  \end{cases}
\vspace{-5pt}
\end{equation}

\noindent{meaning that the agent does not have explicit knowledge of the goal position, and thus experiencing goal reaches is crucial to it forming a representation of state values with respect to the goal.}

To implement our human in-the-loop RL approach above we base it on the Soft Actor Critic (SAC) method \cite{SAC1}. SAC is an off-policy, maximum entropy RL method. Running off-policy allows us to reuse state-action transitions sampled in previous trials when training our networks. This is crucial for our case where fewer interaction steps are feasible. The maximum entropy framework \cite{entropy} adds an entropy maximisation term to our RL objective function for an optimal policy $\pi^*$, which encourages exploration:

\vspace{-10pt}
\begin{equation}\label{eq:2}
\begin{gathered}
J(\pi) = \sum\limits_{t=0}^{T}\E_{\tau}[r(\boldsymbol{s_t},\boldsymbol{a_t}) -
\underbrace{\alpha_{t}log\pi_t(\boldsymbol{a_t} \vert \boldsymbol{s_t})}_\text{Entropy term}]\\
\pi^*=\arg\max_{\pi}J(\pi)
\end{gathered}
\vspace{-3pt}
\end{equation}

\vspace{-3pt}
\noindent{We can balance the exploration/exploitation relationship by the temperature parameter $\alpha$, where a larger $\alpha$ encourages more exploration, and a smaller one corresponds to more exploitation. Using the automatic entropy tuning method introduced by Haarnoja et al. \cite{SAC}, we can constrain the policy's entropy to a desired value throughout the learning process. This removes the need for us to perform intricate hyperparameter tuning, allowing for a very sample-efficient training process.}

Our RL setup learns multiple function approximaters modelled by neural networks. These include the Actor, $\pi(a|s)$, a policy network outputting a Gaussian distribution over the continuous action space. A Critic, $Q(s,a)$, which is represented by two Q-value functions evaluating the value of an action in a given state. Only the minimum of the Q-functions is used for the value gradient, this has shown to speed up training and remove bias in the policy improvement. Additionally, a separate function approximator is used for the soft state value, $V(s)$. All our function approximators above are parametrised with fully connected feed-forward neural networks with two hidden layers, with 32 neurons each, and tanh activations.

During training, the action, $a_t$, is sampled from the Gaussian distribution output of the Actor network. During testing, the mean of the distribution is used, thereby removing the stochasticity, to fully exploit the policy. This, together with the human action ($a_{human}$ in Equation \ref{eq:1}) are applied on the two rotational axes of the game tray. These actions, being velocity commands, are executed on the robot for $200\textrm{ms}$. Limits are set for both the rotational velocity, and the angles of the tray to keep the workspace safe. The resulting state $s_{t+1}$, reward $r_t$ and whether or not the state was terminal, $d$, are then extracted and stored in a replay buffer of past transitions used for gradient updates of the networks. We implement our deep RL agent as described above, using PyTorch \cite{paszke2019pytorch}, and integrate it with the robotic system through ROS.

\subsection{Experimental Setup}
We now have a foundation for applying human and agent actions together on the robot manipulator. Each velocity action is applied constantly for $200\textrm{ms}$. We refer to this as a single control frame. This approach, plus the inherent loop delay and velocity limits of the system, mean that depending on the timing of the human's movements, they  will observe a variable delay of maximum $300\textrm{ms}$ in their intended action being fulfilled. While efforts can be made to reduce this, we see it as an interesting component of the system, as it adds complications from the human's point of view, making the interaction with an untrained RL agent more fair.

For our experimental setup, we define each trial to consist of 200 control frames, a total of 40 seconds. A trial ends immediately if the ball reaches the goal, and otherwise times out in 40 seconds -- i.e. after 200 control frames have been applied. Each trial, therefore, consists of 200 state transitions for the RL agent, which are stored in its replay buffer, to be used for network updates. We set the size of the agent's replay buffer to be 5 trials, meaning a buffer of 1000 (i.e. $5\times200$) state transitions. The game always starts with the ball in one of the corners of the side of the tray opposite the goal-side, alternating between the three corners on each trial. For trial results, scores are defined on a linear scale with a maximum score of 200, and one point lost for each applied control frame -- i.e. if the goal is not reached by the end of the trial, the score will be zero. 




\section{EXPERIMENTS}
\label{sec:experiments}
\subsection{Preliminary study and results}
Pilot experiments of the system were conducted to evaluate its functionality and plan out for the main experiments described in the next sub-section. To closely follow the original application of SAC \cite{haarnoja2018soft}, training is counted in terms of control frames (i.e. RL agent's state transitions), each frame followed by a single gradient update of all networks. All agent transitions are stored in the buffer without limit. Offline updates of the network based on the stored buffer are also performed to accelerate learning.

Two sets of tests were performed in this format. First, a single participant interacted with a previously untrained agent. The training process consisted of 3,500 control frames and 140,000 offline gradient updates. Offline updates were divided throughout training, running 20,000 offline updates for every 500 control frames. After completion of each of these offline updates, performance was tested in trial-based format, for 10 trials, as described before, with results reported averaged over the 10 trials, shown in Figure \ref{pp_results}, left. 

For the second set of tests, 10 participants trained with a previously untrained agent on a trial-based manner. Participants first trained for 8 trials. The agent then undergoes 30,000 offline gradient updates on that buffer followed by a second training set of 7 trials, with transitions added to the original buffer, resulting in a 15-trial long buffer. Another 30,000 offline gradient updates are then applied. Ten trials of testing with the agent follow, with scores averaged and reported. Each participant was then asked to do another ten trials, this time with a human ``expert'' (one of the system's designers who had the most experience with the game) acting as the interaction partner for human-human trials, controlling the axis previously under control of the RL agent. During the human-human trials we ensured the two players cannot see each other, and that they do not communicate in any way. Score results are averaged over ten trials and reported in Figure \ref{pp_results}, right. Results from the single-subject experiment (Figure \ref{pp_results}, left) show the human-agent team as able to solve the interaction task within the time provided. Furthermore, performance consistency is increasing as the human-robot team learn to collaborate. This is possibly the effect of both the agent learning, and human motor adaptation. In the second experiment, the RL-agent's ability to collaborate with humans was compared to how humans collaborate with each other. In five out of the ten preliminary participants (S1, S4, S5, S6, S7) there was no significant differences in performance between the two scenarios. The remaining half of the participants exhibit worse performance when collaborating with the agent. We observed that the players having worse results with their agents, also failed to reach the goal, or at most reached it one time, in the first 500 control frames of the game, which affects the agent's representation of the game's goal. This might be due to these participants being inherently worse players at the game, and would perhaps have been resolved with longer training.

\subsection{Co-learning experiments}
Having confirmed the feasibility of human in-the-loop learning with our system, we move to experiments on collaborative learning. We ran 7 participants. To accelerate learning, participants start their training on a common pre-trained agent. The pre-trained agent is the result of 8 trials of interactions by the expert player from the preliminary study, followed by 30,000 offline gradient updates -- total elapsed time is about 15 minutes. This is effectively half of the training done on the agents in the preliminary study. The pre-trained agent is able to navigate the ball towards the general direction of the goal, with coarse movements and low precision, not making it beyond the obstacle barrier.
\vspace{-1pt}

\begin{figure}[tp]
    \centering
    \includegraphics[width=\columnwidth]{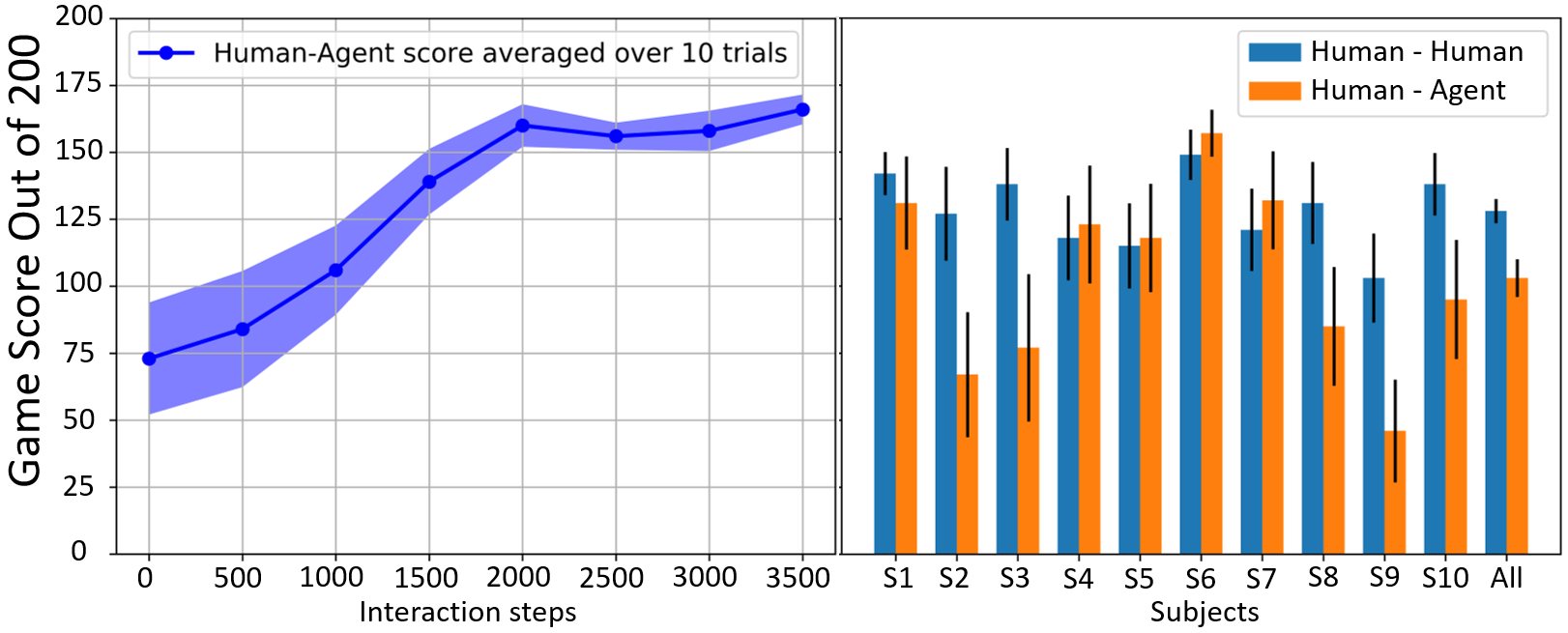}
    \caption{Left: Learning curve of a single human player training with the agent, including both online and offline updates of the agent. Tests are performed at 500-step intervals, scores averaged over ten trials. Plot shows mean score and standard error of the mean. Right: Results of ten participants playing the game with their trained agent, and with a human expert. Mean score and standard error of the mean is shown.}
    \label{pp_results}
\vspace{-5pt}
\end{figure}

Participants are completely naive to the experimental setup. Before training starts, a description of the system is given to the participant. They are told about the RL agent, with a brief description of how RL works. They are also told about what they can control, and how to do it. Each participant is allowed to try out the interface and rotate the tray for 40 seconds. This is without the ball on the tray, and without the RL agent acting. 

Training is done in a trial-based manner, allowing us to observe performance results during training. The replay buffer's length is limited to 5 trials. Training consists of 80 trials, performed in blocks of 10, with the participant given a chance to rest briefly in between blocks. The agent's policy is not updated during the trial. At the end of each trial (i.e. 200 control frames or goal reached), the agent undergoes 200 gradient updates. No further offline gradient updates are performed. Before each trial starts, the participant is alerted by three beeps played over speakers, and a trial's end is similarly announced, by a single beep. Score results and the full state space of the agent's data are recorded for analysis.

Once 80 trials of training are complete, the participants are tested with their own final agent, as well as four agents trained with different players. The agents are frozen during testing, and are not learning any more. Three of the four agents are selected from those of the preliminary study, namely that of S1, S5 and S7 (see Figure \ref{pp_results}, right), which showed a performance at the same level as human-human performance. The fourth agent is the expert player's agent, trained for 160 trials with online updates, followed by 256,000 offline gradient updates on the full buffer of 160 trials. Participants start testing by playing 10 trials with their own agent, then 10 trials each with the four other models (S1, S5, S7 and expert, randomised), and finally playing another set of 10 trials on their own agent. They are not told that their own agent is among the testing agents, but are rather told that they are being tested with 6 unspecified agents. Game score and observed data are recorded for analysis. See supplementary video for better understanding of experiments. 


\section{RESULTS \& DISCUSSION}
\label{sec:results}

\begin{figure}[tp]
    \centering
    \includegraphics[width=\columnwidth]{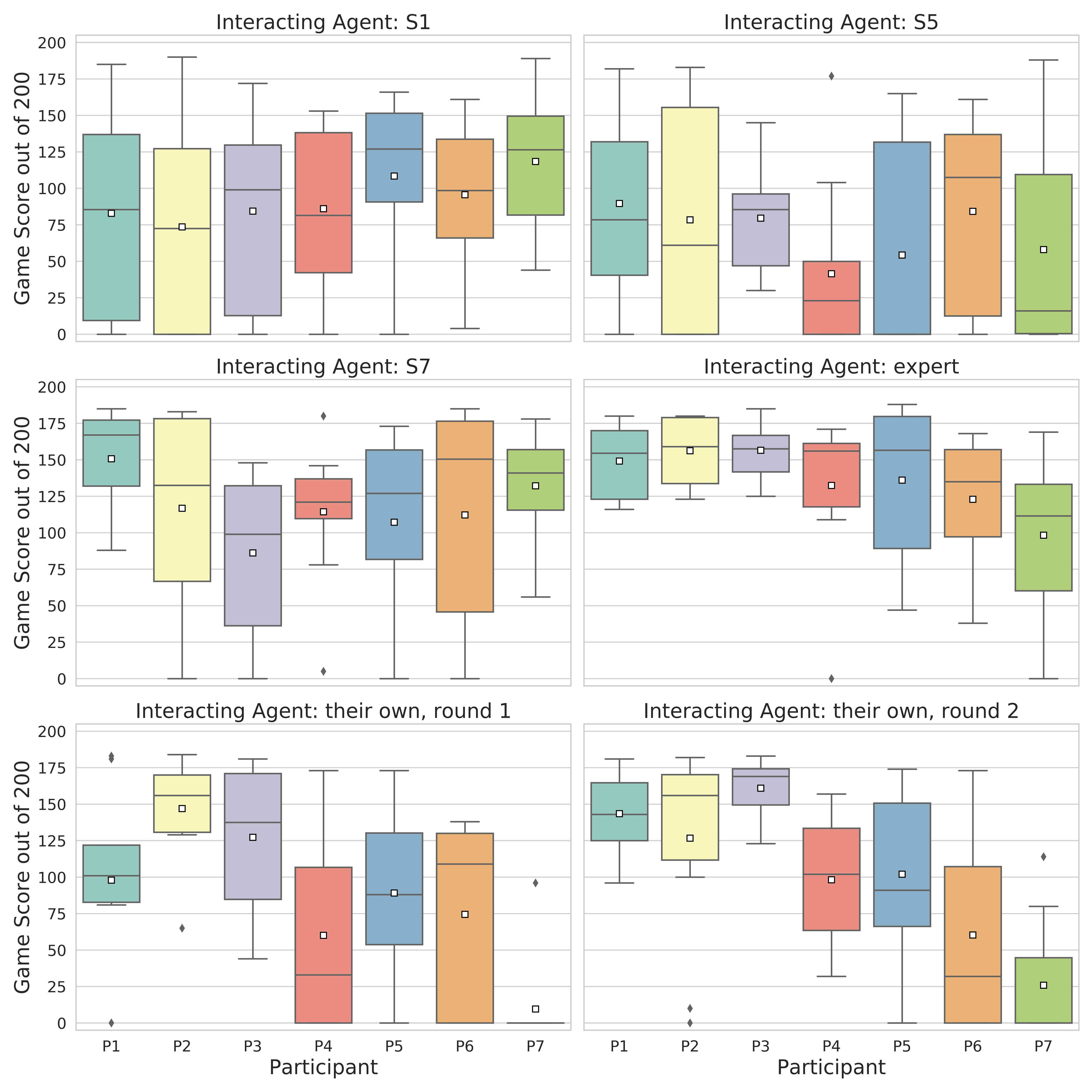}
    \caption{Boxplots of game scores of all 7 participants (P1 to P7) playing with different agents: S1, S5, S7, expert, and their own agent twice. The white squares indicate the mean. }
    \label{boxplot_agents_results}
\vspace{-5pt}
\end{figure}


The results of all 7 players during testing with different agents and their own is shown in Figure \ref{boxplot_agents_results}. We see a divide in the participants' results. Looking at when participants play with their own model, particularly on round 2, we see 3 participants that are performing consistently well (P1, P2, and P3). P4 and P5 have medium performances, whereas the others have bad performances (P6 and P7). This divide seems to persist with some of the other agents, e.g. when playing with the expert agent, we see that, again, P1, P2 and P3 have more consistent performance than the others. This can be explained with P5, P6, and P7 being generally bad at the game – but this does not explain the results when playing with S1. In this case, P1, P2 and P3 show very inconsistent and mediocre performance, significantly lower than their performance with their own agents, whereas P5, P6 and P7 retain their average to high performance that they showed with the expert agent, and outperform their results with their own agents.

This result fits well with the hypothesis that co-learning is occurring, and that personal models are important. P1, P2 and P3 have managed to develop a consistent collaborative policy through their 80 trials, whereas this has occurred less so for P4, and even less so for P5, P6 and almost not at all for P7. However, we can already see from the results that the issue with P5, P6 and P7 is the agent they developed, and not necessarily an inherent skill issue, i.e. there exist agents that improve their game. As an example, see P7's performance with S7, which is on level with the highest performances achieved by any participant with any agent. Perhaps this could have been achieved with their own agent with longer training, or further policy update iterations.

To further analyse this, we compare the different trained agents, independent of the human interacting with them. To do this, we ``test-drive'' our agents offline, by feeding them state iterations, evenly distributed to cover a fair sample of all possible state ranges for all 8 state parameters. We iterate $x$ and $y$ with $5.5\textrm{cm}$ intervals, $\dot{x}$ and $\dot{y}$ with $30 \textrm{cm/s}$ intervals, tray angles along the two axes with $0.05 \textrm{rad}$ intervals and respective angular velocities with $0.2 \textrm{rad/s}$ intervals. Cross-iterating all the state parameters, we record output actions of the agents. This results in an output action vector of length 1,265,625 which can then be used to compare the behaviour of different agents, through correlation analysis.
\begin{figure}[tp]
    \centering
    \includegraphics[width=\columnwidth]{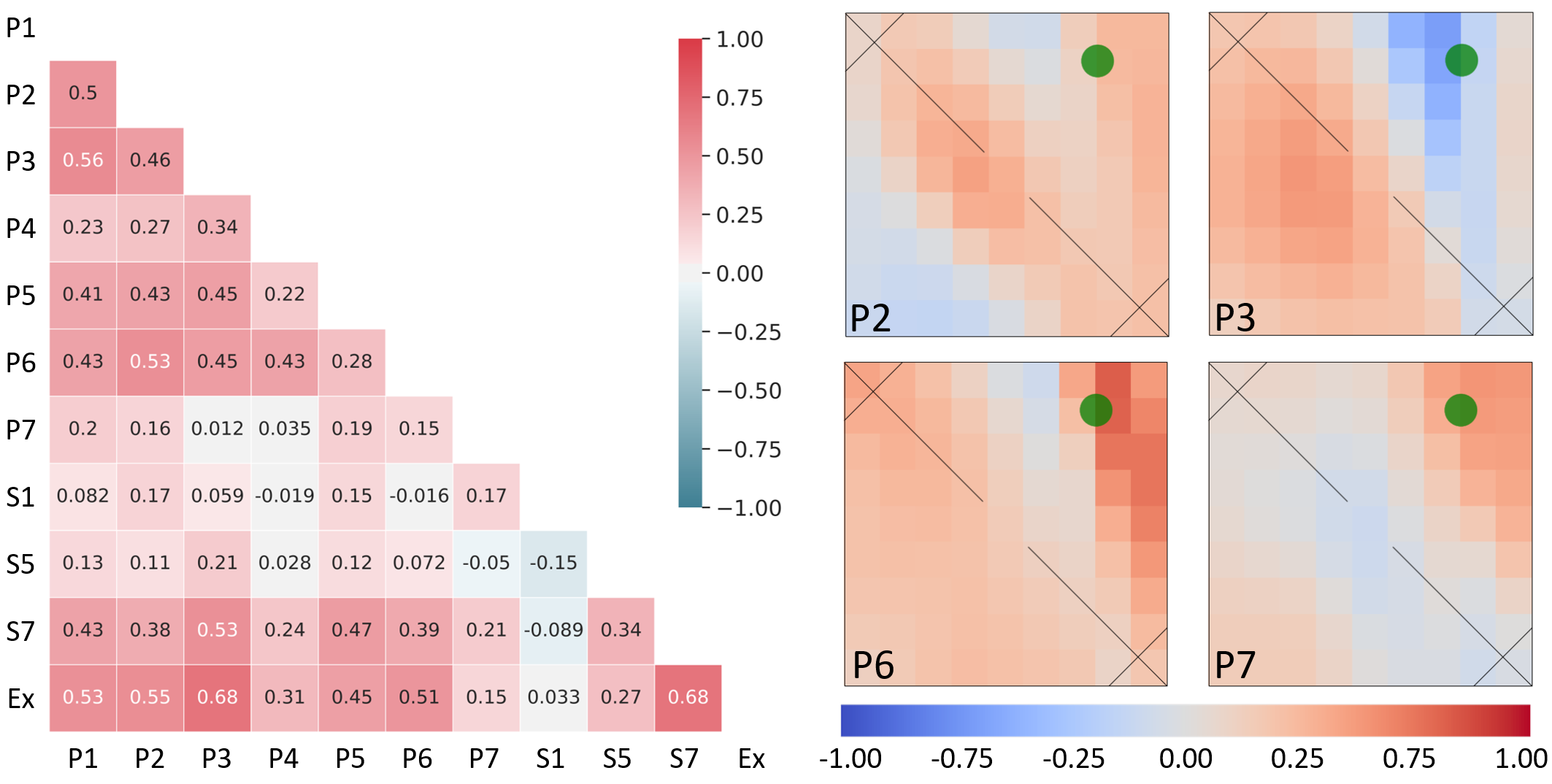}
    \caption{Left: Correlations between the behaviour of all participant-trained agents, as well as the models they tested against, S1, S5, S7 and expert. Right: Spatial representation of trained agents' behaviour correlation with that of the pre-model, for participants P2, P3, P6 and P7. The goal is marked as a green circle -- see Figure \ref{setup_methods} for reference. A higher correlation in a given position means that the final agent's policy has changed less from the original pre-model on which training started.}
    \label{correlations_of_agents_all}
\vspace{-5pt}
\end{figure}
The result of this can be seen in Figure \ref{correlations_of_agents_all}-Left. Note that the expert agent has the highest correlations with the agents of P1, P2 and P3 -- same participants that have the best performance with it. Generally, the expert and S7 agents have the highest correlations with the participants' agents, and they are also the agents that get the best performance from the participants, aside from their own agents - see Figure \ref{boxplot_agents_results}. S1 and S5 have the lowest correlations overall with our participants' agents, and again this fits with the performance plots of Figure \ref{boxplot_agents_results}. The general trend observed by looking at individual participants' agents and how they correlate with test agents, is that the higher a test agent is correlated with the participant's own agent, the better the participant's performance will be with it. Note that the actions of an RL agent in isolation are relating to the behaviour of the person that trained the RL agent, when facing other RL agents. This is an indication that our human-in-the-loop system is leading to co-learning, creating agents that can serve as a representation of the human that trained them, in terms of their skill in this game.


In order to show the meaning of these correlations more intuitively, we present a spatial representation for 4 of the participants across the spectrum. We take P2, P3, P6 and P7. P2 and P3 show generally good performance on their own models, the expert model and S7. P6 and P7 have poor performance overall, though P7 plays well with S7. Figure \ref{correlations_of_agents_all}-Right, shows how these four participants' agents, developed their policies from the pre-model, in a spatial sense. The figure depicts the game tray, with the heatmap values reflecting the correlation of the participant's agent's behaviour in each position, with that of the pre-model on which they started the training. A high correlation means that the pre-model's policy has been retained, whereas lower correlations correspond to higher degrees of change in policy. The pre-model has a good policy around the barrier, and is capable of helping participants get to the goal side of the game. On the goal side however, and particularly the corner closest to the goal, it does not have the best policy: it implements very coarse actions that are hard to coordinate with. We see these reflected in the four participants' policy changes: P2 and P3 have made bigger changes to the policy near the goal, and smaller changes around the barrier, whereas P6 and P7 have done the reverse. This is reflected in their performance results.

\begin{figure}[tp]
    \centering
    \includegraphics[width=\columnwidth]{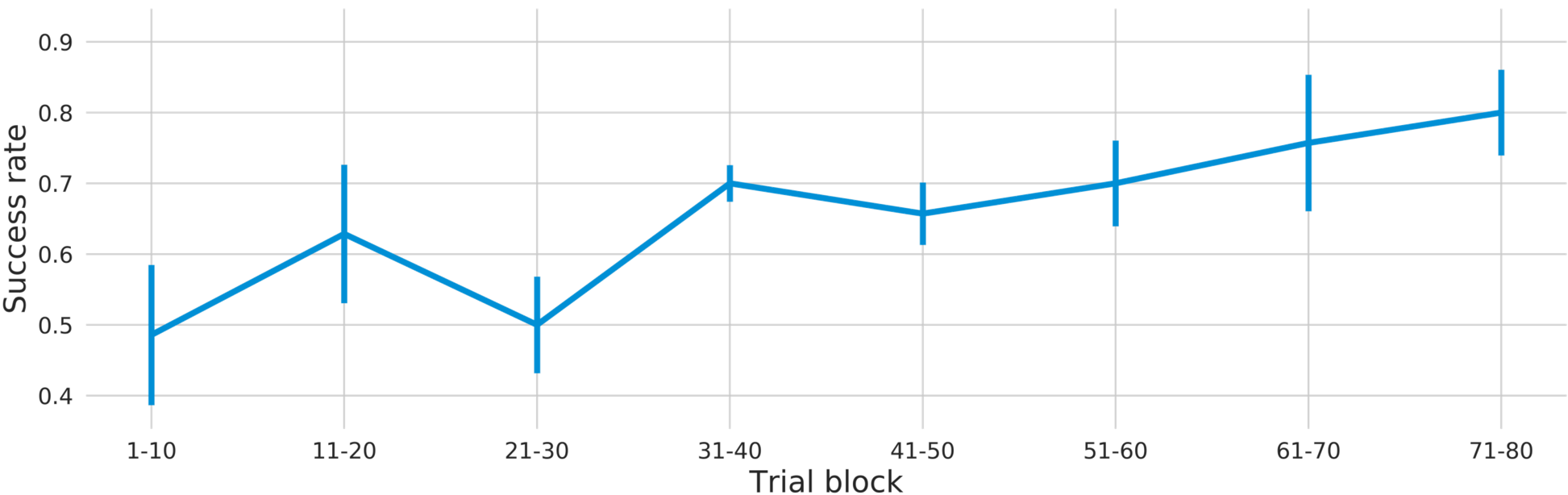}
    \caption{Success rate (reaching the goal) for all 7 participants as they went through 80 trials of training, in 10-trial blocks, serving as the learning curve. Mean and standard error of the mean across all participants shown.}
    \label{results_learning_curve}
\vspace{-5pt}
\end{figure}

Finally, figure \ref{results_learning_curve} is the learning curve based on success rate, i.e. number of times reaching the goal, within 10-trial blocks. Mean values and standard error of the mean across all 7 participants are shown. We see the beginning of a plateau occurring towards the end of the training phase, but further improvement might be possible through further training trials.


\section{CONCLUSIONS} 
\label{sec:conclusions}
We presented a real-world, human in-the-loop, reinforcement learning setup for studies on human-robot collaborative learning. The setup consists of a non-trivial ball and maze game, which can only be solved through effective collaboration. 
Following pilot studies with 10 participants confirming feasibility, we conducted experiments for investigation into human-robot co-learning. We tested 7 participants, for 80 trials. Participants trained on our collaborative motor task together with an RL agent with minimal pre-training. Our results show that with a human in-the-loop it is possible to settle on an effective collaborative policy that leads to consistent success in the game in less than 1 hour of co-training. This is, however, variable across participants, and highly dependent on the particular participant's behaviour during training with the RL agent. This is mirrored in the neuroscience of motor learning, where we have shown evidence of different learning strategies through neural analysis \cite{haar2020brain}. We also see personalisation of RL policies confirmed through analysis on how agents of different participants correlate. Effectively, we are able to relate a human player's performance with new agents to how similar the new agents' policy is to that of their own agent.

We see these outcomes as evidence to the benefits of human in-the-loop systems that augment humans rather than replacing them. Our setup represents a simplified example of a collaborative motor task, but the findings can be extended to real-world applications of human-robot collaboration, particularly in physically sensitive tasks, such as collaborative factory robots \cite{shafti2019real}, assistive robots that restore physical abilities of paralysed users \cite{shafti2019gaze} or augmentation robots that extend our physical capabilities \cite{shafti2020learning}. 



\addtolength{\textheight}{-12cm}   








\bibliographystyle{ieeetr}
\bibliography{references}

\end{document}